\documentclass[11pt,a4paper]{article}
\usepackage{authblk}
\usepackage[hyperref]{acl2018}
\usepackage[utf8]{inputenc}
\usepackage{graphicx}
\usepackage{times}
\usepackage{latexsym}
\usepackage{url}

\aclfinalcopy

\title{Meteorologists and Students: \\A resource for language grounding of geographical descriptors}
 \author[1,2]{Alejandro Ramos-Soto}
 \author[2]{Ehud Reiter}
 \author[2,3]{Kees van Deemter}
 \author[1]{Jose M. Alonso}
 \author[4]{Albert Gatt}
 \affil[1]{\small Centro Singular de Investigación en Tecnoloxías da Información (CiTIUS), Universidade de Santiago de Compostela: \authorcr \tt \small \{alejandro.ramos,josemaria.alonso.moral\}@usc.es}
 \affil[2]{\small Department of Computing Science, University of Aberdeen:
 \authorcr \tt \small \{alejandro.soto,e.reiter,k.vdeemter\}@abdn.ac.uk}
 \affil[3]{\small Utrecht University: \tt \small k.vandeemter@uu.nl}
 \affil[4]{\small Institute of Linguistics and Language Technology, University of Malta: \tt \small albert.gatt@um.edu.mt}

\begin{document}
\maketitle
\begin{abstract}
We present a data resource which can be useful for research purposes on language grounding tasks in the context of geographical referring expression generation. The resource is composed of two data sets that encompass 25 different geographical descriptors and a set of associated graphical representations, drawn as polygons on a map by two groups of human subjects: teenage students and expert meteorologists.
\end{abstract}

\section{Introduction}
Language grounding, i.e., understanding how words and expressions are anchored in data, is one of the initial tasks that are essential for the conception of a data-to-text (D2T) system \cite{bib_connectinglanguage,nlg_datatotext}. This can be achieved through different means, such as using heuristics or machine learning algorithms on an available parallel corpora of text and data \cite{novikova2017e2e} to obtain a mapping between the expressions of interest and the underlying data \cite{bib_mousam}, getting experts to provide these mappings, or running surveys on writers or readers that provide enough data for the application of mapping algorithms \cite{bib_ramos2017empirical}.

Performing language grounding allows ensuring that generated texts include words whose meaning is aligned with what writers understand or what readers would expect \cite{bib_connectinglanguage}, given the variation that is known to exist among writers and readers \cite{reiter2002should}. Moreover, when contradictory data appears in corpora or any other resource that is used to create the data-to-words mapping, creating models that remove inconsistencies can also be a challenging part of language grounding which can influence the development of a successful system \cite{bib_mousam}. 

%To our knowledge, there aren't many resources which are available specifically for language grounding purposes within D2T. These come in the form of a parallel corpus of human-produced texts and data, often meant to feed end-to-end data-driven approaches \cite{novikova2017e2e} or for referring expression generation \cite{gatt2007evaluating}. Corpora can be analyzed to determine how specific words or expressions are actually used by human writers. For example, the SumTime weather forecast corpus was used to analyze the use of certain expressions by meteorologists that were modeled for the SumTime-Mousam system \cite{nlg_mousam}.

This paper presents a resource for language grounding of geographical descriptors. The original purpose of this data collection is the creation of models of geographical descriptors whose meaning is modeled as graded or fuzzy \cite{fisher2000sorites,fisher2006approaches}, to be used for research on generation of geographical referring expressions, e.g.,
\cite{bib_ross_3,nlg_roadsafe,bib_rodrigo,bib_fuzzygre,bib_ramos2017empirical}. However, we believe it can be useful for other related research purposes as well.

\section{The resource and its interest}
\begin{figure}[t]
\centering
\includegraphics[width=\columnwidth]{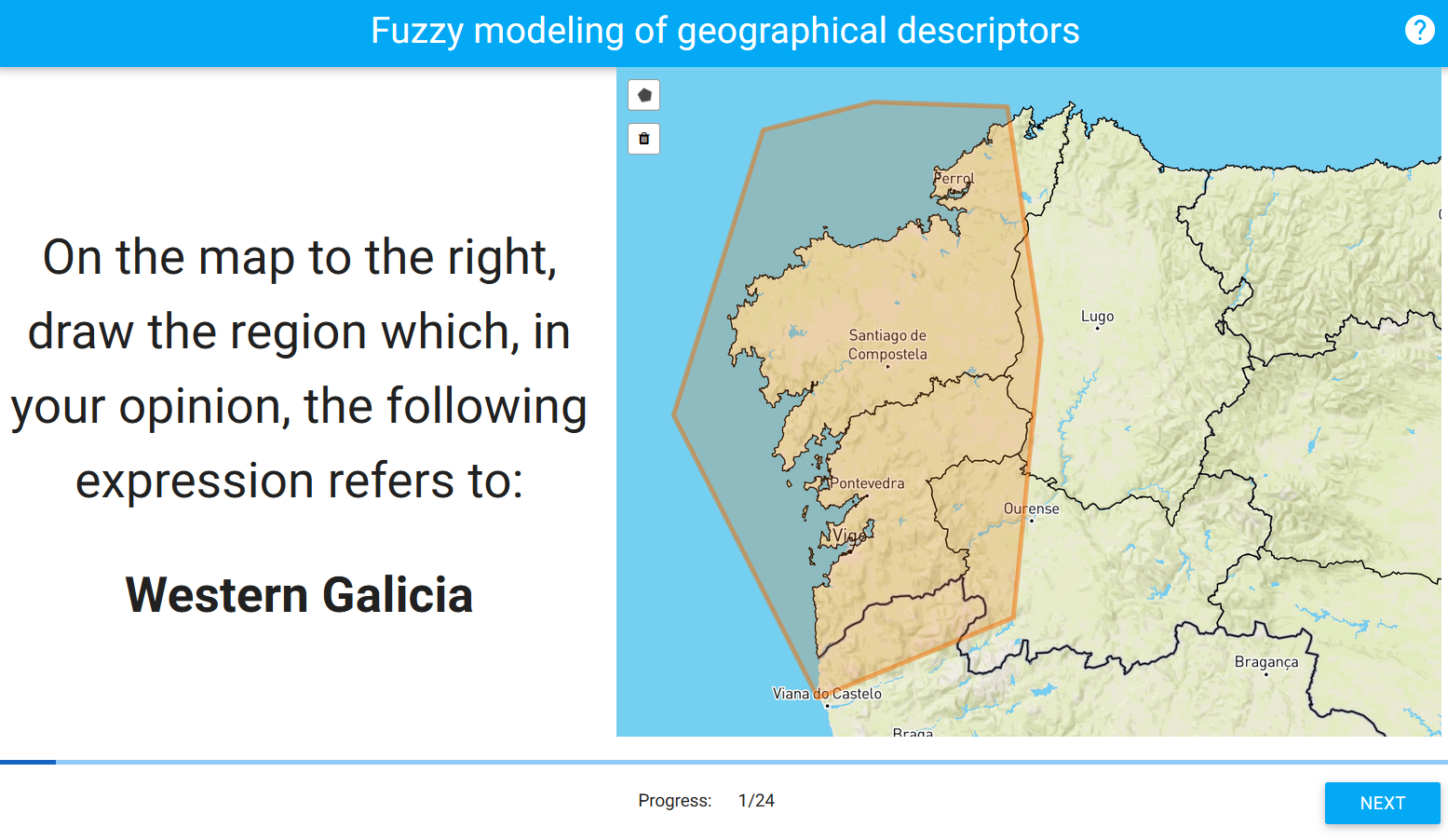}
\caption{Snapshot of the version of the survey answered by the meteorologists (translated from Spanish).}
\label{fig:survey}
\end{figure}

The resource is composed of data from two different surveys. In both surveys subjects were asked to draw on a map (displayed under a Mercator projection) a polygon representing a given geographical descriptor, in the context of the geography of Galicia in Northwestern Spain (see Fig. \ref{fig:survey}). However, the surveys were run with different purposes, and the subject groups that participated in each survey and the list of descriptors provided were accordingly different.

\begin{figure}[b]
\centering
\includegraphics[width=\columnwidth]{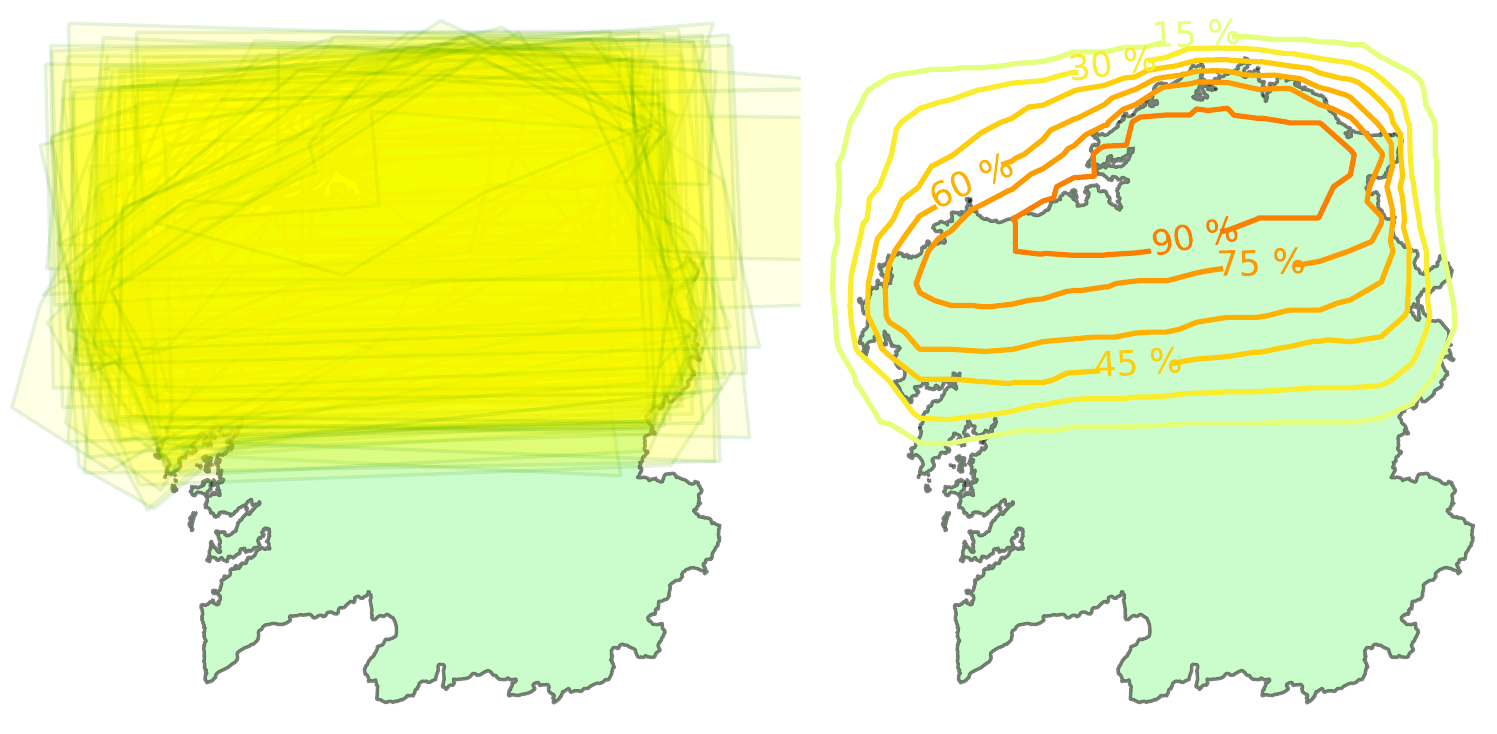}
\caption{Representation of polygon drawings by students and associated contour plot showing the percentage of overlapping answers for ``Northern Galicia''.}
\label{fig:north_teens}
\end{figure}

The first survey was run in order to obtain a high number of responses to be used as an evaluation testbed for modeling algorithms. It was answered by 15/16 year old students in a high school in Pontevedra (located in Western Galicia). 99 students provided answers for a list of 7 descriptors (including cardinal points, coast, inland, and a proper name). Figure \ref{fig:north_teens} shows a representation of the answers given by the students for ``Northern Galicia'' and a contour map that illustrates the percentages of overlapping answers.

\begin{figure}[t]
\centering
\includegraphics[width=\columnwidth]{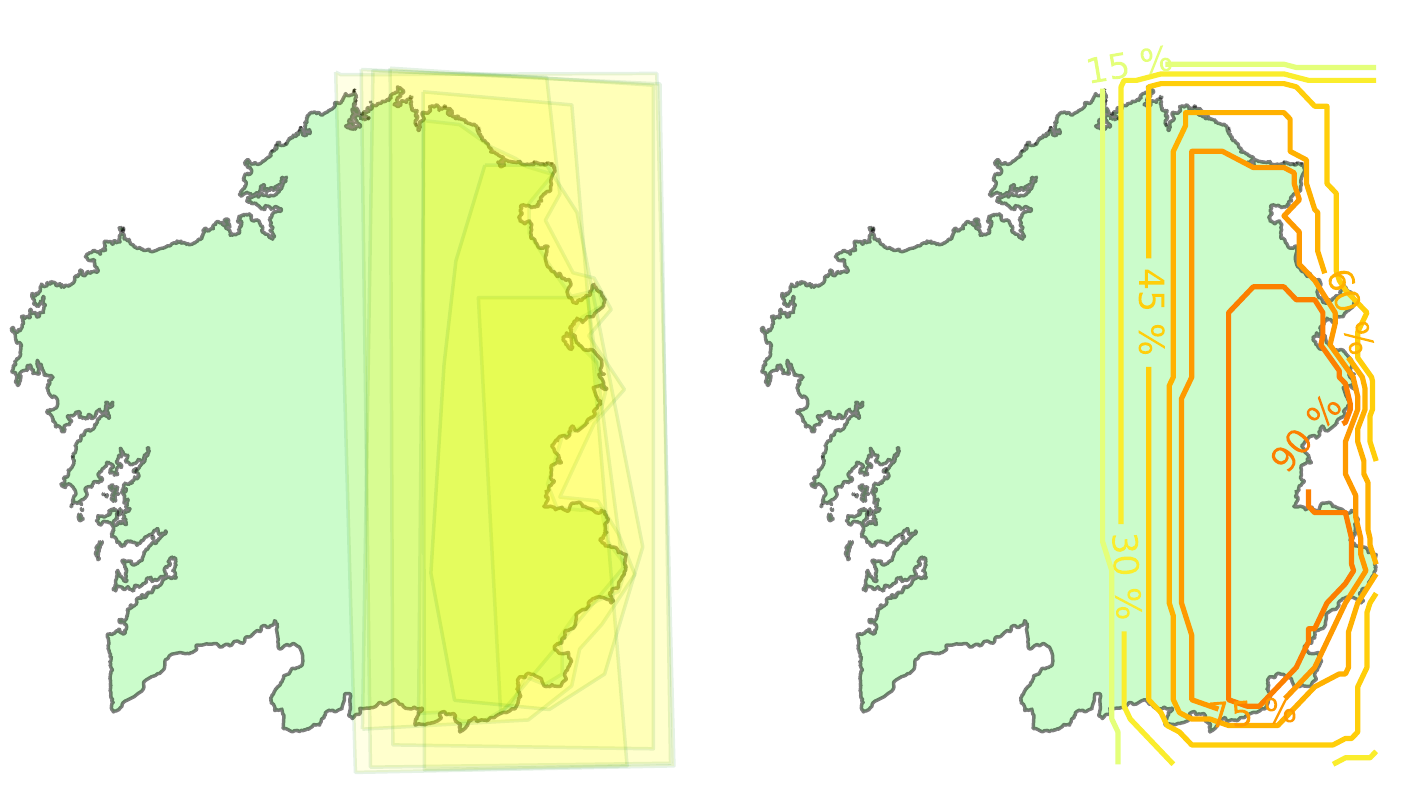}
\caption{Representation of polygon drawings by experts and associated contour plot showing the percentage of overlapping answers for ``Eastern Galicia''.}
\label{fig:east_experts}
\end{figure}

The second survey was addressed to meteorologists in the Galician Weather Agency \cite{meteogalicia}. Its purpose was to gather data to create fuzzy models that will be used in a future NLG system in the weather domain. Eight meteorologists completed the survey, which included a list of 24 descriptors. For instance, Figure \ref{fig:east_experts} shows a representation of the answers given by the meteorologists for ``Eastern Galicia'' and a contour map that illustrates the percentage of overlapping answers.

Table \ref{table:desc_list} includes the complete list of descriptors for both groups of subjects. 20 out of the 24 descriptors are commonly used in the writing of weather forecasts by experts and include cardinal directions, proper names, and other kinds of references such as mountainous areas, parts of provinces, etc. The remaining four were added to study intersecting combinations of cardinal directions (e.g. exploring ways of combining ``north'' and ``west'' for obtaining a model that is similar to ``northwest'').

\begin{table}[b]
\centering
\includegraphics[width=\columnwidth]{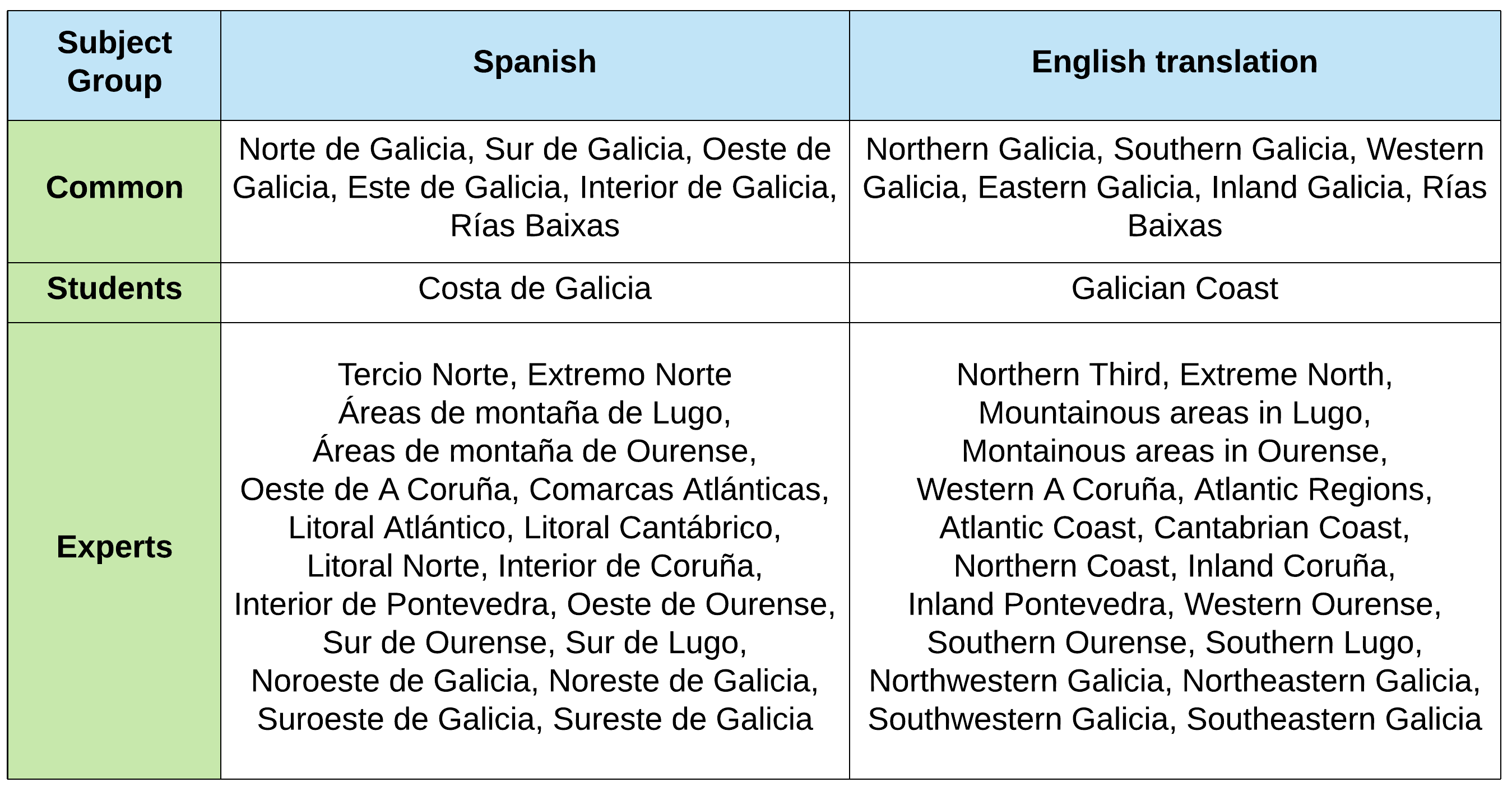}
\caption{List of geographical descriptors in the resource.}
\label{table:desc_list}
\end{table}

The data for the descriptors from the surveys is focused on a very specific geographical context. However, the conjunction of both data sets provides a very interesting resource for performing a variety of more general language grounding-oriented and natural language generation research tasks, such as:
\begin{itemize}
\item Testing algorithms that create geographical models. These models would aggregate the answers from different subjects for each descriptor. The differences among the subjects can be interpreted from a probabilistic or fuzzy perspective that allows a richer characterization of the resulting models. For instance, in Fig. \ref{fig:north_teens} the contour plots could be taken as the basis or support for the semantics of the expression ``Northern Galicia'', with a core region that is accepted by the majority, and a gradual decay as one moves to the outer periphery of the regions outlined.
\item Analyzing differences between the expert and non-expert groups for the descriptors they have in common (as Table \ref{table:desc_list} shows, both groups share 6 descriptors).
\item Studying how to combine models representing the semantics of different cardinal directions, such as ``south'' and ``east'' to obtain a representation of ``southeast''.
\item Developing geographical referring expression generation algorithms based on the empirically created models.
\end{itemize}

\section{Qualitative analysis of the data sets}
The two data sets were gathered for different purposes and only coincide in a few descriptors, so providing a direct comparison is not feasible. However, we can discuss general qualitative insights and a more detailed analysis of the descriptors that both surveys share in common.

At a general level, we had hypothesized that experts would be much more consistent than students, given their professional training and the reduced number of meteorologists participating in the survey. Comparing the visualizations of both data sets we have observed that this is clearly the case; the polygons
drawn by the experts are more concentrated and
therefore there is a higher agreement among them. On top of these differences, some students provided unexpected drawings in terms of shape, size, or location of the polygon for several descriptors.

\begin{figure}[b]
\centering
\includegraphics[width=\columnwidth]{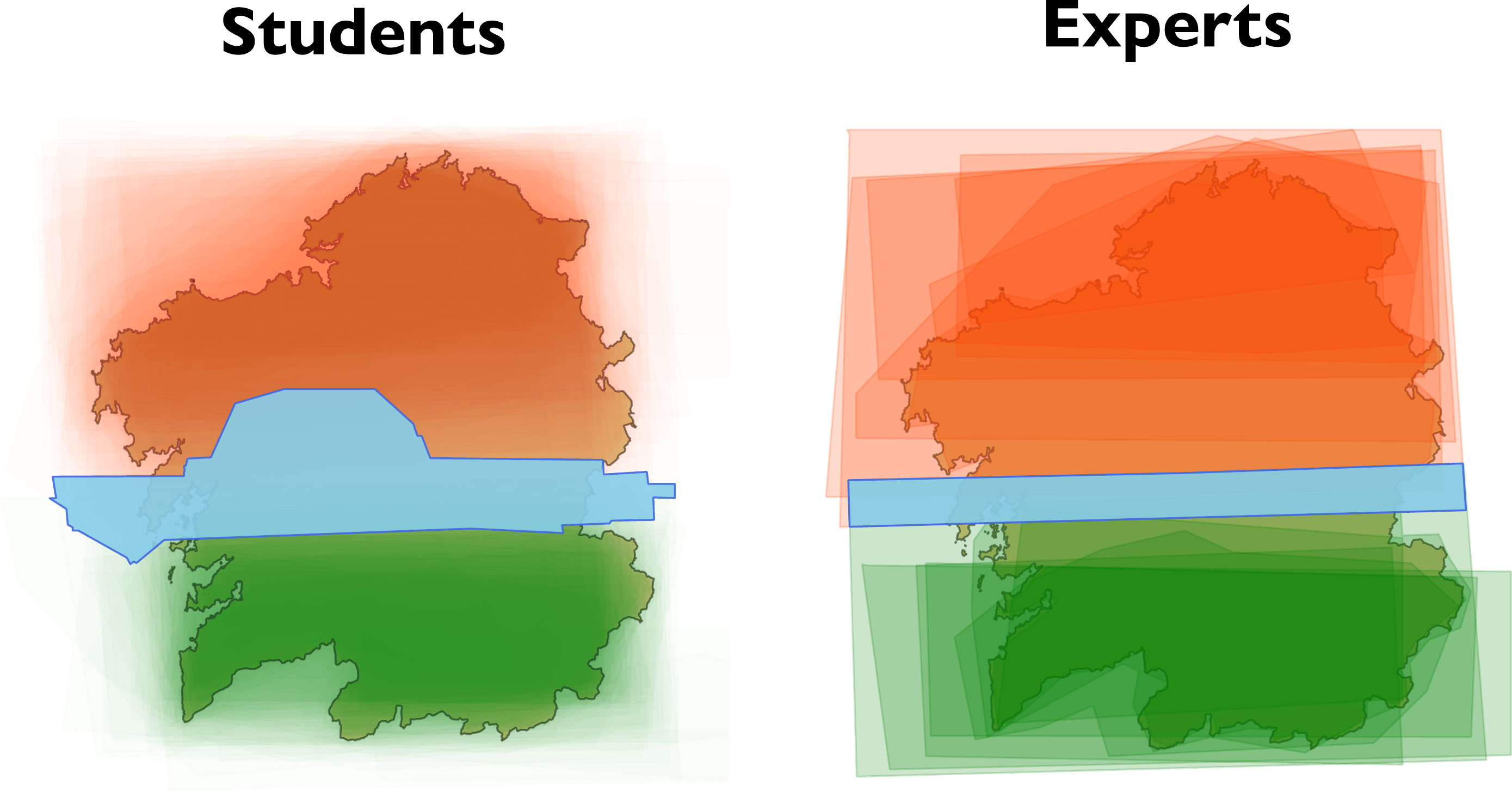}
\caption{Areas overlapping ``north'' and ``south'' for both subject groups (in blue).}
\label{fig:north_south}
\end{figure}

If we focus on single descriptors, one interesting outcome is that some of the answers for ``Northern Galicia'' and ``Southern Galicia'' overlap for both subject groups. Thus, although `north' and `south' are natural antonyms, if we take into account the opinion of each group as a whole, there exists a small area where points can be considered as belonging to both descriptors at the same time (see Fig. \ref{fig:north_south}). In the case of ``west'' and ``east'', the drawings made by the experts were almost divergent and showed no overlapping between those two descriptors.

Regarding ``Inland Galicia'', the unions of the answers for each group occupy approximately the same area with a similar shape, but there is a very high overlapping among the answers of the meteorologists. A similar situation is found for the remaining descriptor ``R\'ias Baixas'', where both groups encompass a similar area. In this case, the students' answers cover a more extensive region and the experts coincide within a more restricted area.
\subsection{A further analysis: apparent issues} \label{sec:issues}
As in any survey that involves a task-based collection of data, some of the answers provided by the subjects for the described data sets can be considered erroneous or misleading due to several reasons. Here we describe for each subject group some of the most relevant issues that any user of this resource  should take into account.

\begin{figure}[b]
\centering
\includegraphics[width=\columnwidth]{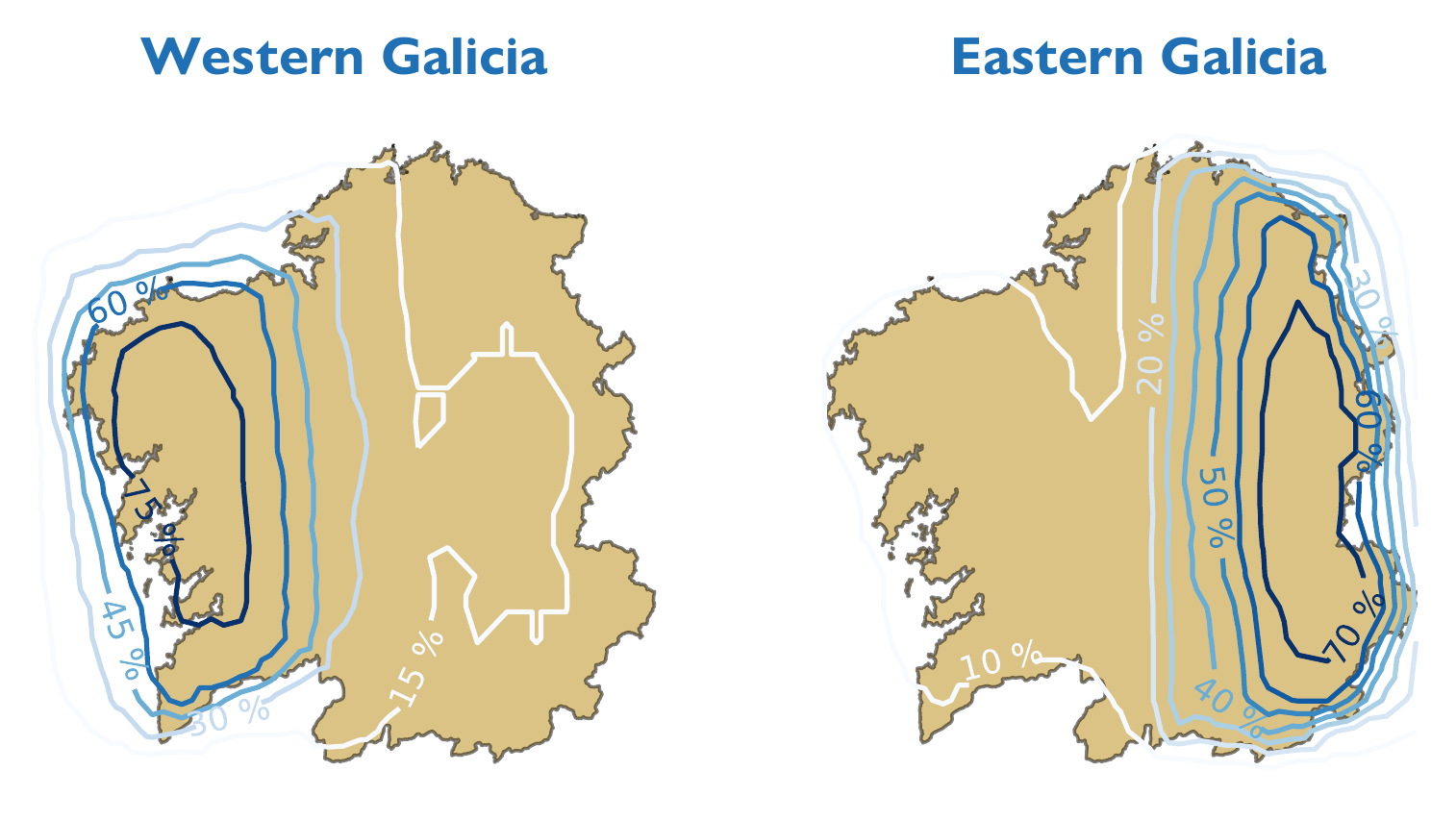}
\caption{Contour maps of student answers for ``Western Galicia'' and ``Eastern Galicia''.}
\label{fig:weast_east_problem}
\end{figure}

In the case of the students, we have identified minor drawing errors appearing in most of the descriptors, which in general shouldn't have a negative impact in the long term thanks to the high number of participants in the original survey. For some descriptors, however, there exist polygons drawn by subjects that clearly deviate from what could be considered a proper answer. The clearest example of this problem involves the `west' and `east' descriptors, which were confused by some of the students who drew them inversely (see Fig. \ref{fig:weast_east_problem}, around 10-15\% of the answers).

In our case, given their background, some of the students may have actually confused the meaning of + ``west'' and ``east''. However, the most plausible explanation is that, unlike in English and other languages, in Spanish both descriptors are phonetically similar (``este'' and ``oeste'') and can be easily mistaken for one another if read without attention.

\begin{figure}[t]
\centering
\includegraphics[width=\columnwidth]{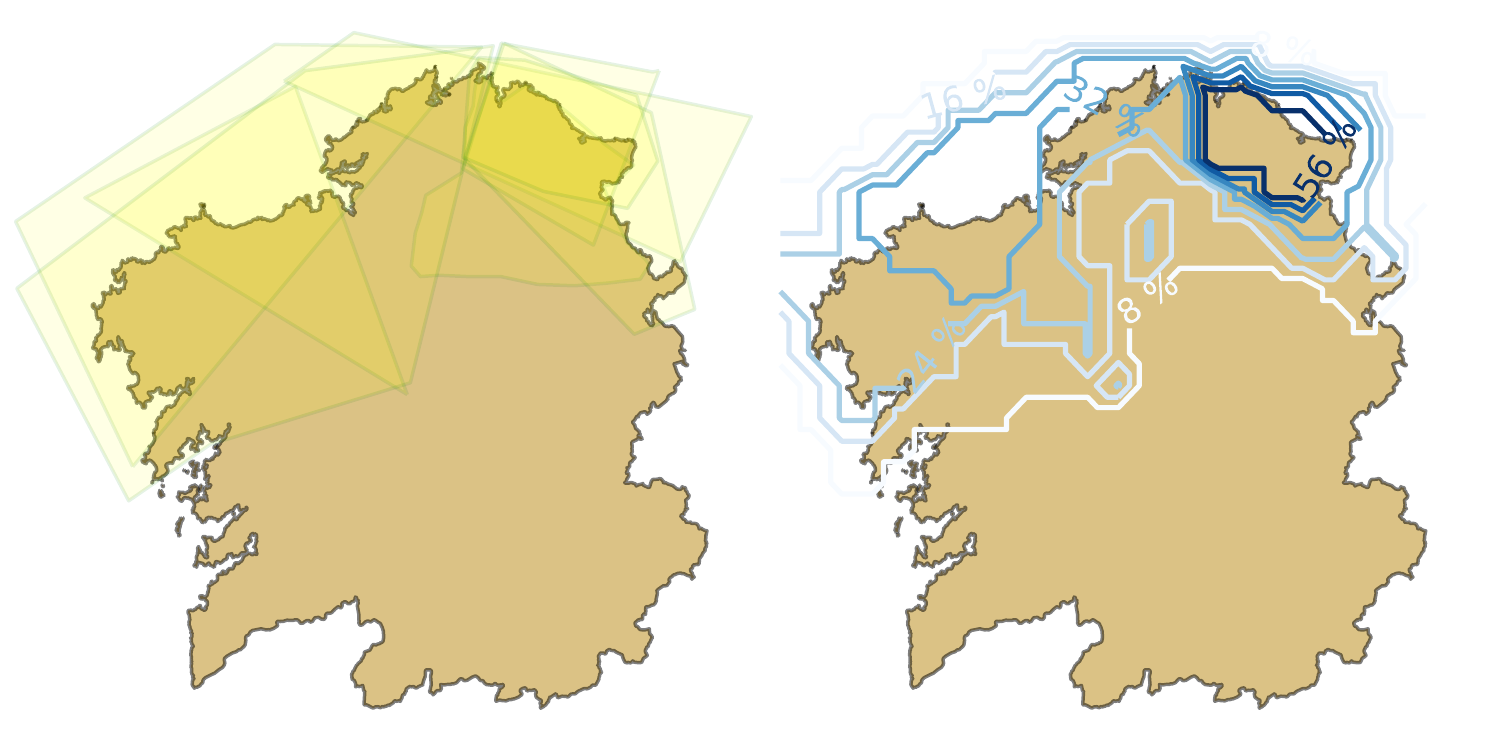}
\caption{Representation of polygon drawings by experts and associated contour plots showing the percentage of overlapping answers for ``Northeastern Galicia''.}
\label{fig:northeast_problem}
\end{figure}

As for the expert group, a similar case is found for ``Northeastern Galicia'' (see Fig. \ref{fig:northeast_problem}), where some of the given answers (3/8) clearly correspond to ``Northwestern Galicia''. However, unlike the issue related to ``west'' and ``east'' found for the student group, this problem is not found reciprocally for the ``northwestern'' answers.

\section{Resource materials}
The resource is available at \cite{bib_resource} under a Creative Commons Attribution-NonCommercial-ShareAlike 4.0 International License. Both data sets are provided as SQLite databases which share the same table structure, and also in a compact JSON format. Polygon data is encoded in GeoJSON format \cite{geojson}. The data sets are well-documented in the repository's README, and several Python scripts are provided for data loading, using Shapely \cite{shapely}; and for visualization purposes, using Cartopy \cite{Cartopy}.

\section{Concluding remarks}
The data sets presented provide a means to perform different research tasks that can be useful from a natural language generation point of view. Among them, we can highlight the creation of models of geographical descriptors, comparing models between both subject groups, studying combinations of models of cardinal directions, and researching on geographical referring expression generation. Furthermore, insights about the semantics of geographical concepts could be inferred under a more thorough analysis.

One of the inconveniences that our data sets present is the appearance of the issues described in Sec. \ref{sec:issues}. It could be necessary to filter some of the answers according to different criteria (e.g., deviation of the centroid location, deviation of size, etc.). For more applied cases, manually filtering can also be an option, but this would require a certain knowledge of the geography of Galicia. In any case, the squared-like shape of this region may allow researchers to become rapidly familiar with many of the descriptors listed in Table \ref{table:desc_list}.

As future work, we believe it would be invaluable to perform similar data gathering tasks for other regions from different parts of the world. These should provide a variety of different shapes (both regular and irregular), so that it can be feasible to generalize (e.g., through data-driven approaches) the semantics of some of the more common descriptors, such as cardinal points, coastal areas, etc. The proposal of a shared task could help achieve this objective.

\section*{Acknowledgments}
 This research was supported by the Spanish Ministry of Economy and Competitiveness (grants  TIN2014-56633-C3-1-R and TIN2017-84796-C2-1-R) and the Galician Ministry of Education (grants GRC2014/030 and "accreditation 2016-2019, ED431G/08"). All grants were co-funded by the European Regional Development Fund (ERDF/FEDER program). A. Ramos-Soto is funded by the ``Conseller\'{i}a de Cultura, Educaci\'{o}n e Ordenaci\'{o}n Universitaria'' (under the Postdoctoral Fellowship accreditation ED481B 2017/030). J.M. Alonso is supported by RYC-2016-19802 (Ram\'on y Cajal contract).
 
 The authors would also like to thank Juan Taboada for providing the list of most frequently used geographical expressions by MeteoGalicia, and Jos\'e Manuel Ramos for organizing the survey at the high school IES Xunqueira I in Pontevedra, Spain.
\bibliography{naaclhlt2016}
\bibliographystyle{acl_natbib}

\end{document}